\documentclass[10pt,conference]{IEEEtran}

\usepackage{cite}
\ifCLASSINFOpdf
   \usepackage[pdftex]{graphicx}
   \graphicspath{{figs/}}
   \DeclareGraphicsExtensions{.pdf,.jpeg,.png}
\else
   \usepackage[dvips]{graphicx}
   \graphicspath{{../figs/}}
   \DeclareGraphicsExtensions{.eps}
\fi
\usepackage[cmex10]{amsmath}
\usepackage{amsthm}

\usepackage{algorithmic}
\usepackage{array}
\ifCLASSOPTIONcompsoc
  \usepackage[caption=false,font=normalsize,labelfont=sf,textfont=sf]{subfig}
\else
  \usepackage[caption=false,font=footnotesize]{subfig}
\fi
\usepackage{balance, multirow, url}
\begin{document}
%
% Copyright Notice
\thispagestyle{empty}
\onecolumn
\linespread{1.2}\selectfont{}
{\noindent\Huge IEEE Copyright Notice}\\[1pt]

{\noindent\large Copyright (c) 2020 IEEE

\noindent Personal use of this material is permitted. Permission from IEEE must be obtained for all other uses, in any current or future media, including reprinting/republishing this material for advertising or promotional purposes, creating new collective works, for resale or redistribution to servers or lists, or reuse of any copyrighted component of this work in other works.}\\[1em]

{\noindent\Large Accepted to be published in: 2020 33rd SIBGRAPI Conference on Graphics, Patterns and Images (SIBGRAPI'20), November 7--10, 2020.}\\[1in]

{\noindent\large Cite as:}\\[1pt]

{\setlength{\fboxrule}{1pt}
 \fbox{\parbox{0.65\textwidth}{S. F. dos Santos and J. Almeida, ``Faster and Accurate Compressed Video Action Recognition Straight from the Frequency Domain,'' in \emph{2020 33rd SIBGRAPI Conference on Graphics, Patterns and Images (SIBGRAPI)}, Recife/Porto de Galinhas, Brazil, 2020, pp. 62-68, doi: 10.1109/SIBGRAPI51738.2020.00017}}}\\[1in]
 
{\noindent\large BibTeX:}\\[1pt]

{\setlength{\fboxrule}{1pt}
 \fbox{\parbox{0.95\textwidth}{
 @InProceedings\{SIBGRAPI\_2020\_Santos,
 
 \begin{tabular}{lll}
  & author    & = \{S. F. \{dos Santos\} and
               J. \{Almeida\}\},\\
			   
  & title     & = \{Faster and Accurate Compressed Video Action Recognition Straight from
               the Frequency Domain\},\\
			   
  & pages     & = \{62--68\},\\
  
  & booktitle & = \{2020 33rd {SIBGRAPI} Conference on Graphics, Patterns and Images ({SIBGRAPI})\},\\
  
  & address   & = \{Recife/Porto de Galinhas, Brazil\},\\
  
  & month     & = \{November 7--10\},\\
  
  & year      & = \{2020\},\\
  
  & publisher & = \{\{IEEE\}\},\\
  
  & doi       & = \{10.1109/SIBGRAPI51738.2020.00017\},\\
  \end{tabular}
  
\}
 }}}

\twocolumn
\linespread{1}\selectfont{}
\clearpage

\title{Faster and Accurate Compressed Video Action Recognition Straight from the Frequency Domain}

%------------------------------------------------------------------------- 
% change the % on next lines to produce the final camera-ready version 
\newif\iffinal
% \finalfalse
\finaltrue
\newcommand{\cmtid}{135}
%------------------------------------------------------------------------- 


\iffinal

\author{
	\IEEEauthorblockN{Samuel Felipe dos Santos and Jurandy Almeida}\\
	\IEEEauthorblockA{%
	  Instituto de Ci\^{e}ncia e Tecnologia\\
	  Universidade Federal de S\~{a}o Paulo -- UNIFESP\\
	  12247-014, S\~{a}o Jos\'{e} dos Campos, SP -- Brazil\\
	  Email: {\small\texttt{\{felipe.samuel, jurandy.almeida\}@unifesp.br}}
	  }
}


%Telephone: (800) 555--1212, Fax: (888) 555--1212}

\else
  \author{Sibgrapi paper ID: \cmtid \\ }
\fi

\maketitle

\begin{abstract}
Human action recognition has become one of the most active field of research in computer vision due to its wide range of applications, like surveillance, medical, industrial environments, smart homes, among others.
Recently, deep learning has been successfully used to learn powerful and interpretable features for recognizing human actions in videos.
Most of the existing deep learning approaches have been designed for processing video information as RGB image sequences. 
For this reason, a preliminary decoding process is required, since video data are often stored in a compressed format.
However, a high computational load and memory usage is demanded for decoding a video.
To overcome this problem, we propose a deep neural network capable of learning straight from compressed video. 
Our approach was evaluated on two public benchmarks, the UCF-101 and HMDB-51 datasets, demonstrating comparable recognition performance to the state-of-the-art methods, with the advantage of running up to 2 times faster in terms of inference speed.
\end{abstract}


\IEEEpeerreviewmaketitle

%========================================================================
\section{Introduction}
\label{sec:intro}
The problem of automatically recognizing human actions in a real-world video has attracted considerable attention from the computer vision community over the past decade. 
This growing interest has been motivated by the wide range of applications, from surveillance, medical, and industrial environments to smart homes~\cite{TIP_2018_Zhang}.

In general, existing solutions rely on a two-step approach: (\textit{i}) extraction and encoding of features, and (\textit{ii}) classification of features into classes~\cite{ARXIV_2016_Kang}.
Traditional methods extract low-level appearance or motion features (e.g., color, texture or optical flow) from space-time interest points detected in a video and use them as input to train classifiers, like support vector machines~(SVM)~\cite{CIARP_2012_Andrade, ICMR_2012_Penatti, MTA_2017_Duta}.
In recent years, the feature extraction and classification steps have been combined into an end-to-end framework using deep learning, where a high-level representation of the raw inputs is obtained by learning a feature hierarchy, in which features from lower levels are composed to form higher level features~\cite{FTML_2009_Bengio}.

Numerous deep learning methods for human action recognition have appeared in the literature~\cite{ARXIV_2016_Kang, FGR_2017_Asadi-Aghbolaghi, IET-CVI_2017_Koohzadi, IVC_2016_Zhu, IVC_2017_Herath, IJCNN_2017_Wu}. 
In most of them, a video is parsed frame by frame with convolutional neural networks (CNNs) designed for images~\cite{MMM_2017_Duta, SIBGRAPI_2018_Duarte}.
Other methods process videos as image sequences using  2D CNNs, 3D CNNs, or recurrent neural networks (RNNs)~\cite{CVPR_2017_Duta, ICMR_2017_Duta, SIBGRAPI_2019_Santos}.

The main limitation of the aforementioned methods is that they have been designed for processing video information as RGB image sequences.
However, most video data available is often stored in a compressed format, like MPEG-4 and H.264.
Therefore, each video must first be decoded into RGB images before being fed to the network, a task demanding high memory and computational cost~\cite{ICIP_2020_Santos}. 

In this paper, we present a deep neural network for human action recognition able to learn straight from compressed video. 
Our network is a two-stream CNN integrating both frequency (i.e., transform coefficients) and temporal (i.e., motion vectors) information, which can be extracted by parsing and entropy decoding the stream of encoded video data.
This enables to save high computational load in full decoding the video stream and thus greatly speed up the processing time. 

Extensive experiments were conducted on two public benchmarks: UCF-101 and HMDB-51.
Results point that our approach can provide comparable or superior effectiveness to existing baselines, but it is much more efficient, since our network has the lowest computational complexity and is the fastest for performing inferences.

The remainder of this paper is organized as follows. 
Section~\ref{sec:mpeg} briefly reviews video compression algorithms.
Section~\ref{sec:relatedwork} introduces some basic concepts of action recognition and discusses related work. 
Section~\ref{sec:approach} describes our network. 
Section~\ref{sec:results} presents the experimental protocol and the results from the comparison of our approach with other methods. Finally, we offer our conclusions and directions for future work in Section~\ref{sec:conclusion}.

%========================================================================
\section{Video Compression}
\label{sec:mpeg}
The objective of video compression is to reduce the spatio-temporal redundancies by using various image transforms and motion compensation~\cite{MTA_2016_Babu}.
Therefore, a lot of superfluous information can be discarded by processing compressed videos.

Many video compression algorithms split video data into three major picture types: intra-coded (I-frames), predicted (P-frames), and bidirectionally predicted (B-frames). 
In a video stream, such pictures are organized into sequences of groups of pictures~(GOPs)~\cite{SIBGRAPI_2019_Santos}.

A GOP must start with an I-frame and can be followed by any number of I and P-frames, which are also known as anchor frames. 
Several B-frames may appear between each pair of consecutive anchor frames~\cite{SIBGRAPI_2019_Santos}. 

Each video frame is divided into a sequence of non-overlapping macroblocks. 
For a video coded in 4:2:0 format, each macroblock consists of six 8x8 pixel blocks: four luminance (Y) blocks and two chrominance (CbCr) blocks. 
Each macroblock is then either intra- or inter-coded~\cite{SIBGRAPI_2019_Santos}. 

An I-frame is completely intra-coded: every 8x8 pixel block in the macroblock is transformed to the frequency domain using the discrete cosine transformation (DCT). 
The 64 DCT coefficients are then quantized (lossy) and entropy (run length and Huffman, lossless) encoded to achieve compression~\cite{SIBGRAPI_2019_Santos}. 

Each P-frame is predictively encoded with reference to its previous anchor frame (the previous I or P-frame). 
For each macroblock in the P-frame, a local region in the anchor frame is searched for a good match in terms of the difference in intensity. 
If a good match is found, the macroblock is represented by a motion vector to the position of the match together with the DCT encoding of the difference (or residue) between the macroblock and its match. 
The DCT coefficients of the residue are quantized and encoded while the motion vector is differentiated and entropy coded (Huffman) with respect to its neighboring motion vector. 
This is usually known as encoding with forward motion compensation. 
Macroblocks encoded by such a process are called as inter-coded macroblocks~\cite{SIBGRAPI_2019_Santos}. 

In order to achieve further compression, B-frames are bidirectionally predictively encoded using forward and/or backward motion compensation with reference to its nearest past and/or future I and/or P-frames~\cite{SIBGRAPI_2019_Santos}. 

The frame number, frame encoding type (I, P or B), the positions and motion vectors of inter-coded macroblocks, the number of intra-coded blocks, and the DC coefficients of each DCT encoded pixel block can be obtained by parsing and entropy (Huffman) decoding video streams. 
These operations take less than 20\% of the computational load in the full video decoding process~\cite{BOOK_1997_Bhaskaran}.

%========================================================================
\section{Basic Concepts and Related Work}
\label{sec:relatedwork}
The term \textit{action}, although its intuitive and rather simple concept, is hard to be defined, since human actions can take various physical forms, extending from the simplest movement of a limb, like the leg movement on a football kick; to the complex joint movement of a group of limbs, like movements of legs, arms, head, and whole body of a soccer player jumping to head the ball on a corner kick~\cite{IVC_2017_Herath}.

A comprehensive review of human action recognition methods can be found in~\cite{ARXIV_2016_Kang, FGR_2017_Asadi-Aghbolaghi, IET-CVI_2017_Koohzadi, IVC_2016_Zhu, IVC_2017_Herath, IJCNN_2017_Wu}.
The main ideas and results from previous work are briefly discussed next. 

Early approaches rely on hand-crafted features and can be grouped into four categories: (1) spatial-temporal volume-based approaches, (2) skeleton-based approaches, (3) trajectory-based approaches, and (4) global approaches~\cite{IVC_2016_Zhu}. 
In general, these methods are built on the pixel-level and carefully designed to deal with challenging issues, such as occlusions and viewpoint changes~\cite{SIBGRAPI_2019_Santos}. 
Despite such approaches may yield good results, their applicability in the real-world is limited, since they are designed by hand and usually require high expertise for domain-expert knowledge~\cite{IJCNN_2017_Wu}.

Lately, thanks to significant advances introduced by deep learning, data-driven features have become a promising alternative in recent approaches, which can be grouped into five categories: (1) learning from video frames, (2) learning from frame transformations, (3) learning from hand-crafted features, (4) three-dimensional CNNs, and (5) hybrid models~\cite{IVC_2016_Zhu}. 
They are able to learn directly from data without needing to incorporate any domain knowledge and to build a high-level representation of the raw inputs by modeling complex functions to map features at different levels of abstraction~\cite{IJCNN_2017_Wu}.

In general, state-of-the-art deep learning approaches have been designed for processing video information as RGB image sequences.
For storage and transmission purposes, video data are usually available in a compressed format (e.g., MPEG-4 and H.264), therefore it is desirable to directly process the compressed video without decoding~\cite{ICIP_2011_Almeida}. 
Unlike RGB pixel values, DCT coefficients and motion vectors from a compressed video provide useful information about its visual content (e.g., appearance changes and motion cues) and can be easily extracted by partial decoding the video stream~\cite{SIBGRAPI_2019_Santos}.
In this way, it is possible to improve not only effectiveness by taking advantage of richer information, but also efficiency by avoiding the full decoding of the video~\cite{CVPR_2018_Wu}.

The use of compressed domain information by deep learning methods is quite recent and has been exploit only by very few works.
The pioneering work of Zhang~et~al.~\cite{CVPR_2016_Zhang, TIP_2018_Zhang} extended the two-stream architecture of Simonyan~and~Zisserman~\cite{NIPS_2014_Simonyan} to use motion vectors instead of optical flow maps in the temporal stream network. However, videos still need to be decoded, since the spatial stream network is fed with RGB images~\cite{SIBGRAPI_2019_Santos}.

The recent work of Wu~et~al.~\cite{CVPR_2018_Wu} introduced a method named Compressed Video Action Recognition~(CoViAR), which extends the Temporal Segment Networks~(TSN) of Wang~et~al.~\cite{ECCV_2016_Wang} to work with video data in compressed form.
For that, RGB images obtained by decoding I-frames and motion features computed from P-frames are provided as input to a multi-stream CNN, with one stream for each input, which are trained separately and then combined by a simple weighted average of their output scores~\cite{SIBGRAPI_2019_Santos}.
Although this approach is efficient, it requires to decode the frequency domain representation (i.e., DCT coefficients) from I-frames to the spatial domain (i.e., RGB pixel values).

To the best of our knowledge, this is the first work performing human action recognition on compressed videos with CNNs designed to operate directly on frequency domain data.

\begin{figure*}[!htb]
    \centering
    \subfloat[CoViAR]{\includegraphics[width=0.45\textwidth]{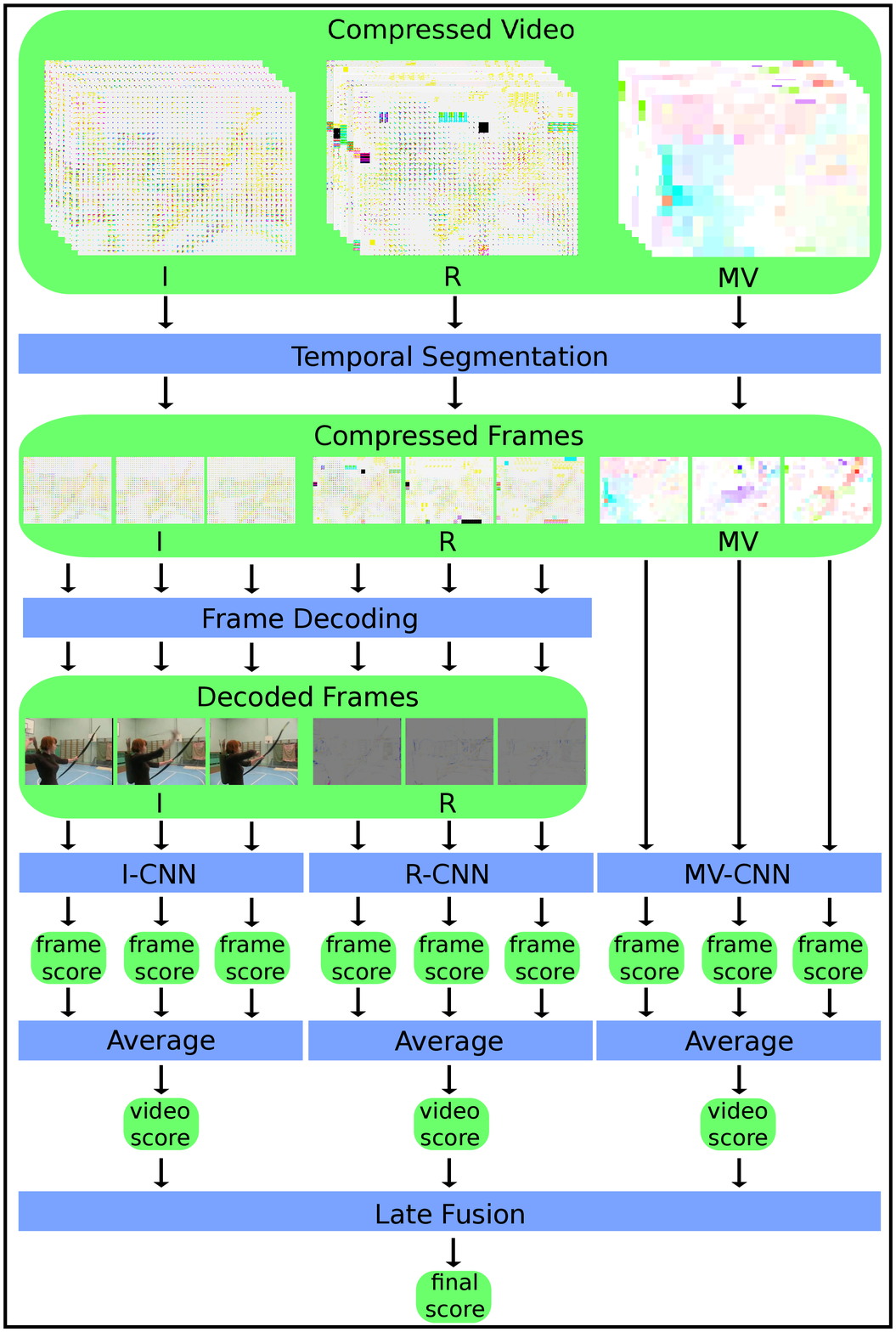}%
    \label{fig:approach:coviar}}
    \hfil
    \subfloat[Fast-CoViAR]{\includegraphics[width=0.45\textwidth]{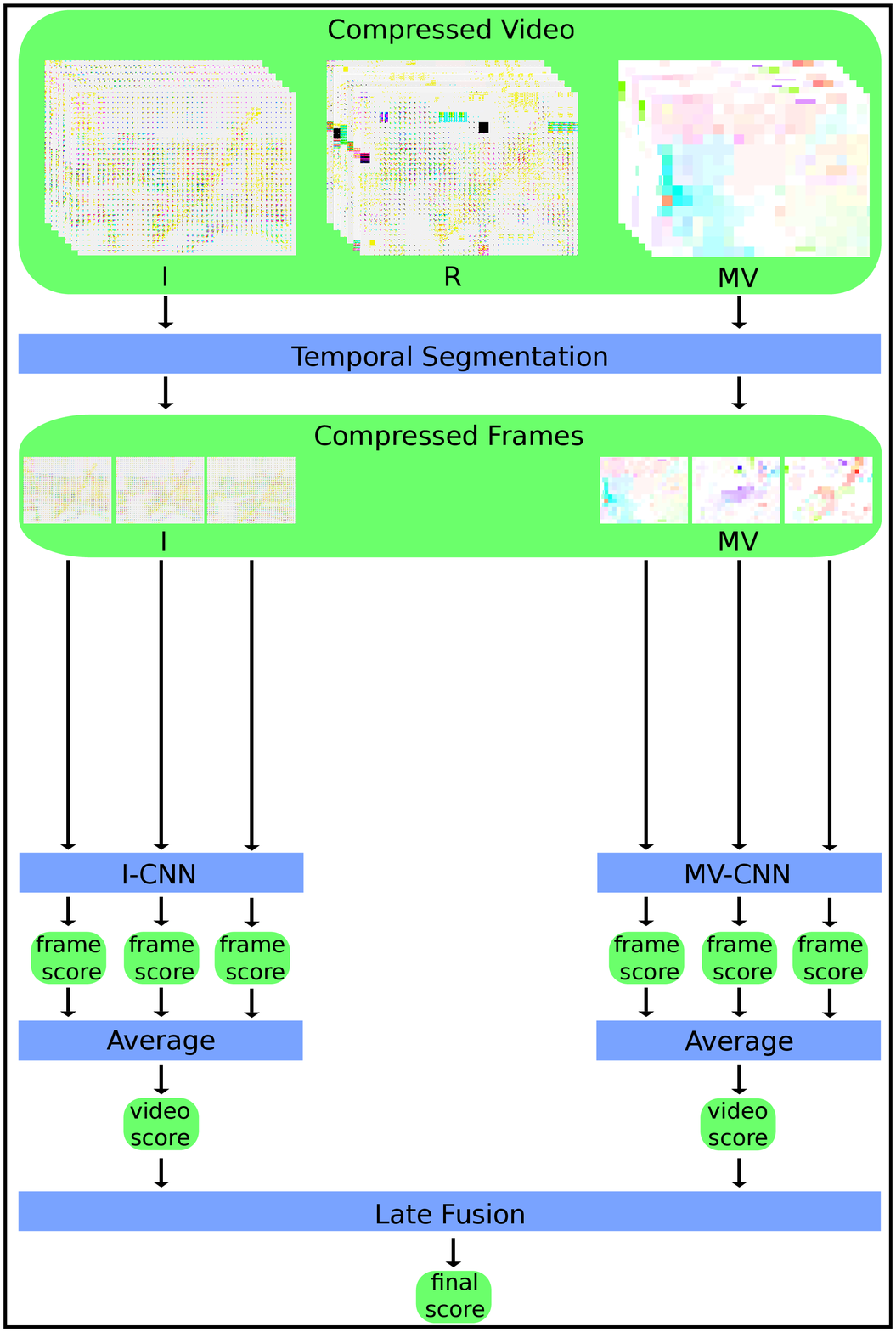}%
    \label{fig:approach:cv-dct}}
    \caption{Illustrations of (a) the CoViAR~\cite{CVPR_2018_Wu} method, and (b) our proposed Fast-CoViAR. Unlike CoViAR, where I-frames need to be decoded before being fed to the network, Fast-CoViAR is designed to operate directly on frequency domain data, learning with DCT coefficients rather than RGB pixels.}
    \label{fig:approach}
\end{figure*}

%========================================================================
\section{Learning from Compressed Videos}
\label{sec:approach}
The starting point for our proposal is the CoViAR~\cite{CVPR_2018_Wu} approach.
In essence, CoViAR extends TSN~\cite{ECCV_2016_Wang} to exploit three information available in MPEG-4 compressed streams: (1) RGB images encoded in I-frames , (2) motion vectors, and (3) residuals encoded in P-frames.
Architecturally, CoViAR is a multi-stream network composed of three independent CNNs, one for each of these three information.
Similar to TSN~\cite{ECCV_2016_Wang}, CoViAR models long-range temporal dynamics by learning from multiple segments of a video. 
For this, uniform sampling is used to take a set of frames. 
Then, frame scores are obtained by feeding the network with one frame at a time. 
Next, a video score is obtained by averaging the frame scores.
Finally, late fusion is performed to take a final prediction, which is computed by the weighted average of the video scores from all the three streams.

\begin{table*}[!htb]
	\centering
	\caption{The hyperparameters used for training the Fast-CoViAR network.}
	\label{tab:parameters}
    \begin{tabular}{l|c|c|c|c}
        \hline
        \hline
        \multirow{2}{*}{\textbf{Hyperparameter}} & \multicolumn{2}{c|}{\textbf{UCF-101}} & \multicolumn{2}{c}{\textbf{HMDB-51}} \\
        \cline{2-5}
        & I & MV & I & MV \\ 
        \hline
        \hline
        \textit{Initial learning rate}  & 0.00015 & 0.005 & 0.0003 & 0.0025 \\ \hline
        \textit{Total number of epochs} & \multicolumn{2}{c|}{510}  &     220 &     360 \\ \hline
        \textit{The step-decay scheduler setting} & \multicolumn{2}{c|}{150, 270, 390} & 55, 110, 165 & 120, 200, 280 \\
        \hline
        \hline
    \end{tabular}
\end{table*}

Although CoViAR has been designed to operate with video data in the compressed domain, it still demands a preliminary decoding step, since the frequency domain representation (i.e., DCT coefficients) used to encode the pictures in I-frames and the residuals in P-frames needs to be decoded to the spatial domain (i.e., RGB pixel values) before being fed to the network.
In fact, CoViAR relies on CNNs designed for processing RGB images, which are not able to learn directly from frequency domain data.

Motivated by the aforesaid observations, we examine ways of integrating frequency domain information into CNNs.
To present date, little work has been done to exploit the DCT representation widely used in compressed data as input for neural networks~\cite{NIPS_2018_Gueguen, ICIP_2020_Santos}.
Our approach is built on top of a modified version of the ResNet-50 architecture~\cite{CVPR_2016_He} presented by Gueguen~et~al.~\cite{NIPS_2018_Gueguen}, which is adapted to facilitate the learning with DCT coefficients rather than RGB pixels.
However, the changes introduced by Gueguen~et~al.~\cite{NIPS_2018_Gueguen} in the ResNet-50 lead to a significant decrease in its computational efficiency.
To alleviate the computational complexity and number of parameters, Santos~et~al.~\cite{ICIP_2020_Santos} extended the modified ResNet-50 network of Gueguen~et~al.~\cite{NIPS_2018_Gueguen} to include a Frequency Band Selection (FBS) technique for selecting the most relevant DCT coefficients before feeding them to the network.

Roughly speaking, our approach extends CoViAR to take advantage of the ResNet-50 network modified by Santos~et~al.~\cite{ICIP_2020_Santos}, enabling it to operate directly on the frequency domain, speeding up the processing time.
For this reason, we named our approach as Fast-CoViAR (Fast Compressed Video Action Recognition).
The similarities and differences of CoViAR and Fast-CoViAR can be observed in Figure~\ref{fig:approach}.
Architecture wise, Fast-CoViAR is a two-stream network which employs two different CNNs as the front-end to learn the frequency and temporal information of a compressed video, respectively.
Unlike CoViAR, instead of a spatial stream using the ResNet-152 network and RGB pixels as input, the frequency stream corresponds to the modified ResNet-50 network of Santos~et~al.~\cite{ICIP_2020_Santos} and is fed with DCT coefficients from I-frames.
Similar to CoViAR, the temporal stream is implemented by a ResNet-18 network fed with motion vectors from P-frames.
Different from CoViAR, we choose not to include a stream for processing residuals from P-frames, once it only results in a slight increase of performance (i.e., gains less than $0.5\%$) at the cost of a significant increase in computational complexity.

The learning procedure of Fast-CoViAR is performed as follows.
Initially, a compressed video is parsed and a set of encoded frames for each stream is obtained by uniform sampling. 
Then, encoded frames are entropy decoded and passed through the network, one frame at a time, generating frame scores. 
Next, frame scores of each stream are averaged to produce a video score. 
Finally, the final prediction is obtained by a simple late fusion, which consists of a weighted average between the video score of both streams.

%========================================================================
\section{Experiments and Results}
\label{sec:results}
For benchmarking purposes, we conducted our experiments on two public datasets containing a large and varied repertoire of different actions~\cite{CVIU_2013_Chaquet}: UCF-101 and HMDB-51.

The UCF-101 dataset\footnote{\url{http://crcv.ucf.edu/data/UCF101.php} (As of July 2020)}~\cite{ARXIV_2012_Soomro} contains 13,320 videos (27 hours) collected from YouTube. 
All videos are in MPEG-4 format (at 25 frames per second and 320$\times$240 resolution), in color and with sound. 
They are categorized into 101 action classes and their duration varies from 1.06 to 71.04 seconds. 
Each of the action classes is divided into 25 groups containing 4-7 videos with common features, like actors and background.
Those videos have large variations in camera motion, object appearance and pose, illumination conditions, etc. 

The HMDB-51 dataset\footnote{\url{http://serre-lab.clps.brown.edu/resource/hmdb-a-large-human-motion-database/} (As of July 2020)}~\cite{ICCV_2011_Kuehne} is composed of 6,766 videos (6 hours) collected from various sources, such as movies and internet sites like YouTube and Google. 
All videos are in MPEG-4 format (at 30 frames per second and with a fixed height of 240 pixels and width ranging from 176 to 592 pixels), in color and no sound. 
They are distributed among 51 action classes with at least 102 videos in each and their duration varies from 0.64 to 35.44 seconds.
Such videos were annotated with information about camera motion, camera viewpoint, video quality, number of actors, visible body parts, etc. 

For evaluation, three training and testing splits are provided with the UCF-101 and HMDB-51 datasets. 
In our experiments, we follow the official evaluation protocol, which consists in evaluating the default training and testing splits separately and reporting the average accuracy over these three splits.

Two data augmentation strategies were applied during training phase: (1) horizontal flipping with 50\% probability and (2) random cropping with scale jittering.
This latter consists of a cropping area whose size is selected at random from different scales (4 scales for the frequency stream: 1, 0.875, 0.75, and 0.66; and 3 scales for the temporal stream: 1, 0.875, and 0.75) and then resized to match the input size requirements of the network (i.e., 28$\times$28 pixels for the frequency stream and 224$\times$224 pixels for the temporal stream).

For training each stream, we follow CoViAR~\cite{CVPR_2018_Wu} and uniformly sample 3 frames from each video to feed the network.
During testing phase, the action category is predicted by taking a uniform sample of 25 frames, each with 5 crops and horizontal flips, totaling 250 frames per video, which are passed through the network independently, with the final prediction being an average of all frame scores.

The ResNet models of both streams were pre-trained on the ImageNet~\cite{IJCV_2015_Russakovsky} dataset and fine-tuned using Adam~\cite{ARXIV_2015_Kingma} with a batch size of 40.
Step-decay was used to reduce the initial learning rate by a factor of 10 after a number of epochs.
The initial learning rates, the total number of epochs, and the step-decay scheduler setting are presented in Table~\ref{tab:parameters}.

\begin{table*}[!htb]
	\centering
	\caption{
	    Classification accuracy (\%) achieved by Fast-CoViAR in the three splits of the UFC-101 and HMDB-51 datasets.
	    In the frequency stream, ResNet-50 was used for processing DCT coefficients from I-frames; whereas ResNet-18 was used for processing motion vectors from P-frames in the temporal stream.
	    Three versions of ResNet-50 with different input depths were tested for the frequency stream.
	    We compare the performance of each model in isolation and also their late fusion (+) by a weighted average of their output scores.
	    The best and the second best results are highlighted in bold and underlining, respectively.}
	\label{tab:accuracy}
    \begin{tabular}{l|c|c|c|cccc}
        \hline
        \hline
        \multirow{2}{*}{\textbf{Dataset}} & \textbf{Picture}  & \textbf{Network} & \textbf{Input} & \multicolumn{4}{c}{\textbf{Accuracy}}\\
        \cline{5-8}
        & \textbf{Type} & \textbf{Architecture} & \textbf{Data} & \textit{Split1} & \textit{Split2} & \textit{Split3} & \textit{Average}\\
        \hline
        \multicolumn{8}{c}{} \\
        \hline
        \multirow{7}{*}{UCF-101} & P   & ResNet-18 & MV    & 67.6 & 68.9 & 70.9 & 69.1 \\
        \cline{2-8}
        % &\multirow{4}{*}{I} & \multirow{4}{*}{ResNet-50} & RGB & 81.7 & 80.6 & 82.1 & 81.5 \\
        %   &&& DCT             & 78.8 & 80.8 & 80.6 & 80.0 \\
        &\multirow{3}{*}{I} & \multirow{3}{*}{ResNet-50} & DCT & 78.8 & 80.8 & 80.6 & 80.0 \\
           &&& DCT w/ FBS (32) & 80.9 & 80.7 & 80.4 & 80.7 \\
           &&& DCT w/ FBS (16) & 78.9 & 76.7 & 78.3 & 78.0 \\
        \cline{2-8}
        % &\multirow{4}{*}{I+P} & \multirow{4}{*}{\parbox{15mm}{\centering ResNet-18\\ +\\ ResNet-50}} & MV + RGB & 87.3 & 86.2 & 87.3 & 86.9 \\
        % &&& MV + DCT              & 84.7 & 85.6 & 86.6 & 85.7 \\
        &\multirow{3}{*}{I+P} & \multirow{3}{*}{\parbox{15mm}{\centering ResNet-18\\ +\\ ResNet-50}} & MV + DCT & 84.7 & \underline{85.6} & \textbf{86.6} & \underline{85.7} \\
        &&& MV + DCT w/ FBS (32)  & \textbf{85.5} & \textbf{85.9} & \underline{86.4} & \textbf{86.0} \\
        &&& MV + DCT w/ FBS (16)  & \underline{85.0} & 83.9 & 85.1 & 84.7 \\
        \hline
        \multicolumn{8}{c}{} \\
        \hline
        \multirow{7}{*}{HMDB-51} & P   & ResNet-18 & MV    & 38.5 & 37.3 & 40.4 & 38.7 \\
        \cline{2-8}
        % &\multirow{4}{*}{I} & \multirow{4}{*}{ResNet-50} & RGB & 47.5 & 46.1 & 44.2 & 45.9 \\
        %   &&& DCT             & 47.8 & 44.6 & 42.4 & 45.0 \\
        &\multirow{3}{*}{I} & \multirow{3}{*}{ResNet-50} & DCT & 47.8 & 44.6 & 42.4 & 45.0 \\
           &&& DCT w/ FBS (32) & 45.9 & 43.6 & 41.9 & 43.8 \\
           &&& DCT w/ FBS (16) & 46.6 & 42.2 & 40.6 & 43.1 \\
        \cline{2-8}
        % &\multirow{4}{*}{I+P} & \multirow{4}{*}{\parbox{15mm}{\centering ResNet-18\\ +\\ ResNet-50}} & MV + RGB & 55.9 & 54.6 & 53.1 & 54.6 \\
        % &&& MV + DCT              &  56.1 & 52.3 & 51.3 & 53.3 \\
        &\multirow{3}{*}{I+P} & \multirow{3}{*}{\parbox{15mm}{\centering ResNet-18\\ +\\ ResNet-50}} & MV + DCT & \textbf{56.1} & \textbf{52.3} & \textbf{51.3} & \textbf{53.3} \\
        &&& MV + DCT w/ FBS (32)  &  \underline{55.8} & \underline{51.5} & \underline{50.9} & \underline{52.7} \\
        &&& MV + DCT w/ FBS (16)  &  55.7 & 50.4 & 50.3 & 52.1 \\
        \hline
        \hline
    \end{tabular}
\end{table*}

\begin{table}[!htb]
	\centering
	\caption{Computational complexity (GFLOPs) and number of parameters of the original ResNet-50 with RGB inputs and its modified versions using DCT~\cite{ICIP_2020_Santos}.}
	\label{tab:complexity}
    \begin{tabular}{ccccc}
        \hline
        \hline
        \textbf{Network} & \textbf{Input Data} & \textbf{Channels} & \textbf{GFLOPs} & \textbf{Params} \\
        \hline
        \multirow{4}{*}{ResNet-50} & RGB               & 3$\times$1  & 3.86 & 25.6M \\
         & DCT               & 3$\times$64 & 5.40 & 28.4M \\
         & DCT with FBS (32) & 3$\times$32 & 3.68 & 26.2M \\
         & DCT with FBS (16) & 3$\times$16 & 3.18 & 25.6M \\
        \hline
        \hline
    \end{tabular}
\end{table}

Three different versions of the ResNet-50 network were tested for processing I-frames in the frequency stream: the modified version of Gueguen~et~al.~\cite{NIPS_2018_Gueguen}, which has an input depth of 64 DCT coefficients for each color channel and is referred as DCT; and its two improved versions proposed by Santos~et~al.~\cite{ICIP_2020_Santos} and denoted as DCT w/ FBS, which uses the FBS technique to select 32 and 16 DCT coefficients for each color channel as input to the network, respectively.
The computational complexity, measured by the amount of floating point operations (FLOPs) required during a forward pass through the neural network, and the number of parameters reported by Santos~et~al.~\cite{ICIP_2020_Santos} for the original ResNet-50 network and its modified versions are presented in Table~\ref{tab:complexity}. 

The experiments were performed on a machine equipped with two 10-core Intel Xeon E5-2630v4 2.2 GHz processors, 64 GBytes of DDR4-memory, and 1 NVIDIA Titan Xp GPU. 
The machine runs Linux Mint 18.1 (kernel 4.4.0) and the ext4 file system. 
Fast-CoViAR was implemented in PyTorch (version 1.2.0) upon the CoViAR implementation\footnote{\url{https://github.com/chaoyuaw/pytorch-coviar} (As of July 2020)}.

Table~\ref{tab:accuracy} presents the classification accuracy achieved by Fast-CoViAR in each of the three splits of the UFC-101 and HMDB-51 datasets. 
We compare the results obtained by each stream in isolation and also for their combination.
Among the different versions of ResNet-50 tested for the frequency stream, for the UCF-101 dataset, the one with the best performance was DCT, followed closely by DCT w/ FBS (32); while for the HMDB-51 dataset, the best was DCT w/ FBS (32) followed by DCT.
This shows that the FBS technique is beneficial for the network, reducing its computational complexity without sacrificing accuracy.
Notice that the frequency stream performs better than the temporal stream, however the results are improved when they are combined, showing that motion vectors from P-frames offer complementary information to DCT coefficients from I-frames.
The best results were achieved by combining these two inputs, reaching classification accuracies of 86.0\% on the UCF-101 dataset and 53.3\% on the HMDB-51 dataset.
These results indicate that the use of the information readily available in a compressed video is promising. 

Table~\ref{tab:complexity2} compares the computational complexity and classification accuracy of Fast-CoViAR and CoViAR.
Also, we considered the results reported by Wu~et~al.~\cite{CVPR_2018_Wu} for four baselines: ResNet-50~\cite{CVPR_2017_Feichtenhofer}, ResNet-152~\cite{CVPR_2017_Feichtenhofer}, C3D~\cite{ICCV_2015_Tran}, and Res3D~\cite{ARXIV_2017_Tran}.
Similar to CoViAR~\cite{CVPR_2018_Wu}, the computational costs for processing I-frames and P-frames are different and, for this reason, the values reported for Fast-CoViAR are the average GFLOPs over all frames, 
In terms of classification accuracy, Fast-CoViAR achieved the second best performance on the UCF-101 dataset and the third best performance on the HMDB-51 dataset.
The highest classification accuracies were achieved by CoViAR.
Since CoViAR and Fast-CoViAR use the same network for processing motion vectors from P-frames and the accuracy gains obtained by CoViAR with residuals from P-frames are marginal (less than 0.5\%), we believe that it is because CoViAR processes information from I-frames using ResNet-152~\cite{CVPR_2016_He}, which is much deeper than ResNet-50 used by Fast-CoViAR.
However, Fast-CoViAR has the smallest network computation complexity among all the baselines and is up to 2 times faster than CoViAR.

\begin{table}[!htb]
	\centering
    \caption{Comparison of the computation complexity (GFLOPs) and classification accuracy (\%) of different networks. The best and the second best results are highlighted in bold and underlining, respectively.}
    \label{tab:complexity2}
    \begin{tabular}{l|c|c|c|c}
        \hline
        \hline
        \multicolumn{2}{c|}{\multirow{2}{*}{\textbf{Approach}}} & \multirow{2}{*}{\textbf{GFLOPs}} & \multicolumn{2}{c}{\textbf{Accuracy (\%)}} \\
        \cline{4-5}
        \multicolumn{2}{l|}{} & & \textit{UCF-101} & \textit{HMDB-51} \\
        \hline
        \multicolumn{5}{c}{} \\
        \hline
        \multicolumn{2}{l|}{ResNet-50~\cite{CVPR_2017_Feichtenhofer}}  &  3.8 & 82.3 & 48.9 \\
        \multicolumn{2}{l|}{ResNet-152~\cite{CVPR_2017_Feichtenhofer}} & 11.3 & 83.4 & 46.7 \\
        \multicolumn{2}{l|}{C3D~\cite{ICCV_2015_Tran}}                 & 38.5 & 82.3 & 51.6 \\
        \multicolumn{2}{l|}{Res3D~\cite{ARXIV_2017_Tran}}              & 19.3 & 85.8 & \underline{54.9} \\
        \multicolumn{2}{l|}{CoViAR~\cite{CVPR_2018_Wu}$^4$}            &  4.2 & \textbf{90.4} & \textbf{59.1} \\
        %CV-C3D~\cite{SIBGRAPI_2019_Santos}        &  \textbf{0.5} & 83.9 & \underline{55.7} \\
        \hline
        \multicolumn{5}{c}{} \\
        \hline
        % \textbf{CoViAR~\cite{CVPR_2018_Wu} I+MN w/ ResNet-50} & 2.3   & \underline{86.9} & 54.6 \\
        \multirow{3}{*}{Fast-CoViAR} & DCT & 2.7             & 85.7  & 53.3 \\
        & DCT w/ FBS (32)                  & \underline{2.3} & \underline{86.0}  & 52.7 \\
        & DCT w/ FBS (16)                  & \textbf{2.1}    & 84.7  & 52.1 \\
        \hline
        \hline
    \end{tabular}
    \\[1em]
    \parbox{0.44\textwidth}{\footnotesize{$^4$For a fair comparison, we considered the results reported by CoViAR~\cite{CVPR_2018_Wu} using only information from compressed domain. To improve its accuracy, CoViAR use optical flow besides motion vectors.}}
\end{table}

Table~\ref{tab:comparison} compares the classification accuracy of Fast-CoViAR and the state-of-the-art compressed video methods.
In general, Fast-CoViAR obtained comparable results on both the UCF-101 and HMDB-51 datasets, showing that it retains high accuracy while greatly reducing computational cost.
Despite the results for EMV-CNN and DTMV-CNN were slightly better than Fast-CoViAR, in addition to motion vectors, they also use optical flow during the training phase. 
This feature can also be used by Fast-CoViAR, but its computation is significantly slower since video decoding is required.

\begin{table}[!htb]
	\centering
    \caption{Comparison of the classification accuracy (\%) on the UCF-101 and HMDB-51 datasets for state-of-the-art compressed video based methods. The best and the second best results are highlighted in bold and underlining, respectively.}
    \label{tab:comparison}
    \begin{tabular}{l|c|c|c}
        \hline
        \hline
        \multicolumn{2}{c|}{\textbf{Approach}} & \textbf{UCF-101} & \textbf{HMDB-51} \\
        \hline
        \multicolumn{4}{c}{} \\
        \hline
        \multicolumn{2}{l|}{EMV-CNN~\cite{CVPR_2016_Zhang}}     & 86.4 & 51.2$^5$ \\
        \multicolumn{2}{l|}{DTMV-CNN~\cite{TIP_2018_Zhang}}     & \underline{87.5} & \underline{55.3} \\
        \multicolumn{2}{l|}{CoViAR~\cite{CVPR_2018_Wu}}         & \textbf{90.4}    & \textbf{59.1} \\
        \hline
        %CV-C3D~\cite{SIBGRAPI_2019_Santos} & 83.9  & \underline{55.7} \\
        \multicolumn{4}{c}{} \\
        \hline
        %\textbf{CoViAR~\cite{CVPR_2018_Wu} I+MN w/ ResNet-50}  & 86.93 & 54.55 \\
        \multirow{3}{*}{Fast-CoViAR}      & DCT                      & 85.7 & 53.3 \\
                                          & DCT w/ FBS(32)           & 86.0 & 52.7 \\
                                          & DCT w/ FBS(16)           & 84.7 & 52.1 \\
        \hline
        \hline
    \end{tabular}
    \\[1em]
    \parbox{0.42\textwidth}{\footnotesize{$^5$This result was reported in~\cite{TIP_2018_Zhang} and refers to the classification accuracy obtained only on Split 1 of the HMDB-51 dataset. We included here just for reference.}}    
\end{table}

The computational efficiency is the key advantage of Fast-CoViAR. 
To evaluate its efficiency, we measured the average inference time per-frame, which refers to the time spent for a forward pass through the network.
For this, we sum up the total time taken to feed all the streams sequentially. 
To obtain a fair comparison, the forwarding time of CoViAR was measured using the authors' implementation, upon which we implemented Fast-CoViAR using the same code optimization.

\begin{figure}[!htb]
    \centering
    \subfloat[UCF-101]{\includegraphics[width=0.5\textwidth]{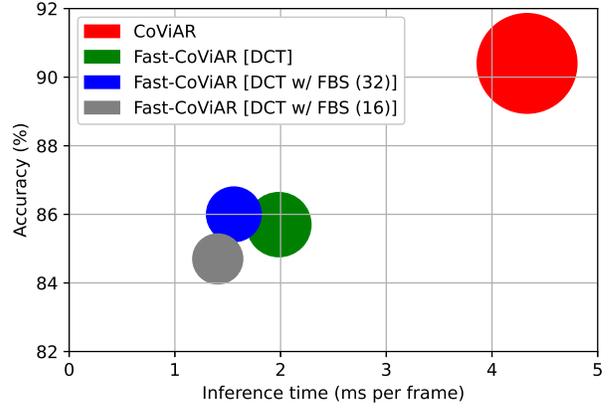}%
    \label{fig:inference:ucf101}} 
    \\
    \subfloat[HMDB-51]{\includegraphics[width=0.5\textwidth]{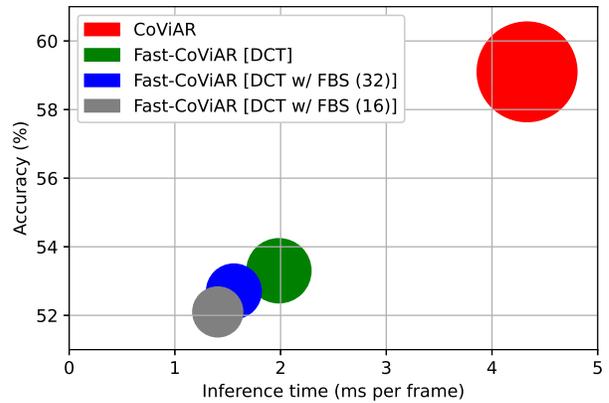}%
    \label{fig:inference:hmdb51}}
    \caption{
    Comparison of the classification accuracy (\%) and the inference time (ms per frame) for Fast-CoViAR and CoViAR on the UCF-101 and HMDB-51 datasets. Node size denotes the network computation complexity (GFLOPs).}
    \label{fig:inference}
\end{figure}

Figure~\ref{fig:inference} compares the classification accuracy, the network computation complexity, and the inference time for Fast-CoViAR and CoViAR on the UCF-101 and HMDB-51 datasets.
Although CoViAR yields a higher classification accuracy, Fast-CoViAR leads to a significant speed-up, being around twice faster for inferences.
Among the three variations of Fast-CoViAR, DCT w/ FBS (32) performs similar to or better than DCT, but is much more faster.

%========================================================================
\section{Conclusion}
\label{sec:conclusion}
In this paper, we proposed a novel deep neural network for human action recognition.
Our network takes advantage of the information readily available in a compressed video, saving high computational load in full decoding the video stream and greatly speeding up the processing time.

The novelty of our approach is that it was designed to operate directly on frequency domain data, learning with DCT coefficients rather than RGB pixels.
Its architecture consists of a two-stream CNN integrating both frequency (i.e., transform coefficients) and temporal (i.e., motion vectors) information, whose predictions are combined by late fusion.

To validate our approach, we conducted experiments on the UCF-101 and HMDB-51 datasets.
The results demonstrated that the recognition performance of our network is similar to the state-of-the-art methods and its inference speed is up to 2 times faster.
In short, our network is both faster and accurate.

As future work, we plan to evaluate the use of 3D network architectures in our approach, like Res3D~\cite{ARXIV_2017_Tran} or I3D~\cite{CVPR_2017_Carreira}.
Also, we want to evaluate smarter fusion strategies to combine the predictions of the two streams and to search for the optimal parameters of our network.
In addition, we intend to evaluate our approach in large-scale datasets, like Kinetics~\cite{ARXIV_2017_Kay}, and in other applications besides action recognition.


\iffinal
\section*{Acknowledgment}
This research was supported by the FAPESP-Microsoft Research Virtual Institute (grant~2017/25908-6) and the Brazilian National Council for Scientific and Technological Development - CNPq (grants~423228/2016-1 and 313122/2017-2). 
We gratefully acknowledge the support of NVIDIA Corporation with the donation of the Titan Xp GPU used for this research.
\fi


\balance
\bibliographystyle{IEEEtran}
% \bibliography{SIBGRAPI_2020_Santos}

\begin{thebibliography}{10}
\providecommand{\url}[1]{#1}
\csname url@samestyle\endcsname
\providecommand{\newblock}{\relax}
\providecommand{\bibinfo}[2]{#2}
\providecommand{\BIBentrySTDinterwordspacing}{\spaceskip=0pt\relax}
\providecommand{\BIBentryALTinterwordstretchfactor}{4}
\providecommand{\BIBentryALTinterwordspacing}{\spaceskip=\fontdimen2\font plus
\BIBentryALTinterwordstretchfactor\fontdimen3\font minus
  \fontdimen4\font\relax}
\providecommand{\BIBforeignlanguage}[2]{{%
\expandafter\ifx\csname l@#1\endcsname\relax
\typeout{** WARNING: IEEEtran.bst: No hyphenation pattern has been}%
\typeout{** loaded for the language `#1'. Using the pattern for}%
\typeout{** the default language instead.}%
\else
\language=\csname l@#1\endcsname
\fi
#2}}
\providecommand{\BIBdecl}{\relax}
\BIBdecl

\bibitem{TIP_2018_Zhang}
B.~Zhang, L.~Wang, Z.~Wang, Y.~Qiao, and H.~Wang, ``Real-time action
  recognition with deeply transferred motion vector cnns,'' \emph{{IEEE}
  Transactions on Image Processing}, vol.~27, no.~5, pp. 2326--2339, 2018.

\bibitem{ARXIV_2016_Kang}
S.-M. Kang and R.~P. Wildes, ``Review of action recognition and detection
  methods,'' \emph{CoRR}, vol. abs/1610.06906, 2016.

\bibitem{CIARP_2012_Andrade}
F.~S.~P. Andrade, J.~Almeida, H.~Pedrini, and R.~S. Torres, ``Fusion of local
  and global descriptors for content-based image and video retrieval,'' in
  \emph{Iberoamerican Congress on Pattern Recognition (CIARP'12)}, 2012, pp.
  845--853.

\bibitem{ICMR_2012_Penatti}
O.~A.~B. Penatti, L.~T. Li, J.~Almeida, and R.~S. Torres, ``A visual approach
  for video geocoding using bag-of-scenes,'' in \emph{{ACM} International
  Conference on Multimedia Retrieval (ICMR'12)}, 2012, pp. 1--8.

\bibitem{MTA_2017_Duta}
I.~C. Duta, J.~R.~R. Uijlings, B.~Ionescu, K.~Aizawa, A.~G. Hauptmann, and
  N.~Sebe, ``Efficient human action recognition using histograms of motion
  gradients and {VLAD} with descriptor shape information,'' \emph{Multimedia
  Tools and Applications}, vol.~76, no.~21, pp. 22\,445--22\,472, 2017.

\bibitem{FTML_2009_Bengio}
Y.~Bengio, ``Learning deep architectures for ai,'' \emph{Foundations and Trends
  in Machine Learning}, vol.~2, no.~1, pp. 1--127, 2009.

\bibitem{FGR_2017_Asadi-Aghbolaghi}
M.~Asadi{-}Aghbolaghi, A.~Clap{\'{e}}s, M.~Bellantonio, H.~J. Escalante,
  V.~Ponce{-}L{\'{o}}pez, X.~Bar{\'{o}}, I.~Guyon, S.~Kasaei, and S.~Escalera,
  ``A survey on deep learning based approaches for action and gesture
  recognition in image sequences,'' in \emph{{IEEE} International Conference on
  Automatic Face {\&} Gesture Recognition (FG'17)}, 2017, pp. 476--483.

\bibitem{IET-CVI_2017_Koohzadi}
M.~Koohzadi and N.~M. Charkari, ``Survey on deep learning methods in human
  action recognition,'' \emph{{IET} Computer Vision}, vol.~11, no.~8, pp.
  623--632, 2017.

\bibitem{IVC_2016_Zhu}
F.~Zhu, L.~Shao, J.~Xie, and Y.~Fang, ``From handcrafted to learned
  representations for human action recognition: {A} survey,'' \emph{Image and
  Vision Computing}, vol.~55, pp. 42--52, 2016.

\bibitem{IVC_2017_Herath}
S.~Herath, M.~T. Harandi, and F.~Porikli, ``Going deeper into action
  recognition: {A} survey,'' \emph{Image and Vision Computing}, vol.~60, pp.
  4--21, 2017.

\bibitem{IJCNN_2017_Wu}
D.~Wu, N.~Sharma, and M.~Blumenstein, ``Recent advances in video-based human
  action recognition using deep learning: {A} review,'' in \emph{International
  Joint Conference on Neural Networks (IJCNN'17)}, 2017, pp. 2865--2872.

\bibitem{MMM_2017_Duta}
I.~C. Duta, B.~Ionescu, K.~Aizawa, and N.~Sebe, ``Spatio-temporal {VLAD}
  encoding for human action recognition in videos,'' in \emph{International
  Conference on MultiMedia Modeling (MMM'17)}, 2017, pp. 365--378.

\bibitem{SIBGRAPI_2018_Duarte}
L.~A. Duarte, O.~A.~B. Penatti, and J.~Almeida, ``Bag of attributes for video
  event retrieval,'' in \emph{{SIBGRAPI} -- Conference on Graphics, Patterns
  and Images (SIBGRAPI'18)}, 2018, pp. 447--454.

\bibitem{CVPR_2017_Duta}
I.~C. Duta, B.~Ionescu, K.~Aizawa, and N.~Sebe, ``Spatio-temporal vector of
  locally max pooled features for action recognition in videos,'' in
  \emph{{IEEE} International Conference on Computer Vision and Pattern
  Recognition (CVPR'17)}, 2017, pp. 3205--3214.

\bibitem{ICMR_2017_Duta}
I.~C. Duta, B.~Ionescu, K.~Aizawa, and N.~Sebe, ``Simple, efficient and
  effective encodings of local deep features for video action recognition,'' in
  \emph{{ACM} International Conference on Multimedia Retrieval (ICMR'17)},
  2017, pp. 218--225.

\bibitem{SIBGRAPI_2019_Santos}
S.~F. Santos, N.~Sebe, and J.~Almeida, ``{CV-C3D:} action recognition on
  compressed videos with convolutional 3d networks,'' in \emph{{SIBGRAPI} --
  Conference on Graphics, Patterns and Images (SIBGRAPI'19)}, 2019, pp. 24--30.

\bibitem{ICIP_2020_Santos}
S.~F. Santos, N.~Sebe, and J.~Almeida, ``The good, the bad, and the ugly:
  Neural networks straight from jpeg,'' in \emph{{IEEE} International
  Conference on Image Processing (ICIP'20)}, 2020, pp. 1--5.

\bibitem{MTA_2016_Babu}
R.~V. Babu, M.~Tom, and P.~Wadekar, ``A survey on compressed domain video
  analysis techniques,'' \emph{Multimedia Tools and Applications}, vol.~75,
  no.~2, pp. 1043--1078, 2016.

\bibitem{BOOK_1997_Bhaskaran}
V.~Bhaskaran and K.~Konstantinides, \emph{Image and Video Compression
  Standards: Algorithms and Architectures}, 2nd~ed.\hskip 1em plus 0.5em minus
  0.4em\relax Norwell, MA, USA: Kluwer Academic Publishers, 1997.

\bibitem{ICIP_2011_Almeida}
J.~Almeida, N.~J. Leite, and R.~S. Torres, ``Comparison of video sequences with
  histograms of motion patterns,'' in \emph{{IEEE} International Conference on
  Image Processing (ICIP'11)}, 2011, pp. 3673--3676.

\bibitem{CVPR_2018_Wu}
C.-Y. Wu, M.~Zaheer, H.~Hu, R.~Manmatha, A.~J. Smola, and
  P.~Kr\"{a}henb\"{u}hl, ``Compressed video action recognition,'' in
  \emph{{IEEE} International Conference on Computer Vision and Pattern
  Recognition (CVPR'18)}, 2018, pp. 6026--6035.

\bibitem{CVPR_2016_Zhang}
B.~Zhang, L.~Wang, Z.~Wang, Y.~Qiao, and H.~Wang, ``Real-time action
  recognition with enhanced motion vector cnns,'' in \emph{{IEEE} International
  Conference on Computer Vision and Pattern Recognition (CVPR'16)}, 2016, pp.
  2718--2726.

\bibitem{NIPS_2014_Simonyan}
K.~Simonyan and A.~Zisserman, ``Two-stream convolutional networks for action
  recognition in videos,'' in \emph{Annual Conference on Neural Information
  Processing Systems (NIPS'14)}, 2014, pp. 568--576.

\bibitem{ECCV_2016_Wang}
L.~Wang, Y.~Xiong, Z.~Wang, Y.~Qiao, D.~Lin, X.~Tang, and L.~V. Gool,
  ``Temporal segment networks: Towards good practices for deep action
  recognition,'' in \emph{European Conference on Computer Vision (ECCV'16)},
  2016, pp. 20--36.

\bibitem{NIPS_2018_Gueguen}
L.~Gueguen, A.~Sergeev, B.~Kadlec, R.~Liu, and J.~Yosinski, ``Faster neural
  networks straight from {JPEG},'' in \emph{Annual Conference on Neural
  Information Processing Systems (NIPS'18)}, 2018, pp. 3937--3948.

\bibitem{CVPR_2016_He}
K.~He, X.~Zhang, S.~Ren, and J.~Sun, ``Deep residual learning for image
  recognition,'' in \emph{{IEEE} International Conference on Computer Vision
  and Pattern Recognition (CVPR'16)}, 2016, pp. 770--778.

\bibitem{CVIU_2013_Chaquet}
J.~M. Chaquet, E.~J. Carmona, and A.~Fern{\'{a}}ndez{-}Caballero, ``A survey of
  video datasets for human action and activity recognition,'' \emph{Computer
  Vision and Image Understanding}, vol. 117, no.~6, pp. 633--659, Jun. 2013.

\bibitem{ARXIV_2012_Soomro}
K.~Soomro, A.~R. Zamir, and M.~Shah, ``{UCF101:} {A} dataset of 101 human
  actions classes from videos in the wild,'' \emph{CoRR}, vol. abs/1212.0402,
  2012.

\bibitem{ICCV_2011_Kuehne}
H.~Kuehne, H.~Jhuang, E.~Garrote, T.~A. Poggio, and T.~Serre, ``{HMDB:} {A}
  large video database for human motion recognition,'' in \emph{{IEEE}
  International Conference on Computer Vision (ICCV'11)}, 2011, pp. 2556--2563.

\bibitem{IJCV_2015_Russakovsky}
O.~Russakovsky, J.~Deng, H.~Su, J.~Krause, S.~Satheesh, S.~Ma, Z.~Huang,
  A.~Karpathy, A.~Khosla, M.~S. Bernstein, A.~C. Berg, and F.-F. Li, ``Imagenet
  large scale visual recognition challenge,'' \emph{International Journal of
  Computer Vision}, vol. 115, no.~3, pp. 211--252, 2015.

\bibitem{ARXIV_2015_Kingma}
D.~P. Kingma and J.~Ba, ``Adam: {A} method for stochastic optimization,''
  \emph{CoRR}, vol. abs/1412.6980, 2015.

\bibitem{CVPR_2017_Feichtenhofer}
C.~Feichtenhofer, A.~Pinz, and R.~P. Wildes, ``Spatiotemporal multiplier
  networks for video action recognition,'' in \emph{{IEEE} International
  Conference on Computer Vision and Pattern Recognition (CVPR'17)}, 2017, pp.
  7445--7454.

\bibitem{ICCV_2015_Tran}
D.~Tran, L.~D. Bourdev, R.~Fergus, L.~Torresani, and M.~Paluri, ``Learning
  spatiotemporal features with 3d convolutional networks,'' in \emph{{IEEE}
  International Conference on Computer Vision (ICCV'15)}, 2015, pp. 4489--4497.

\bibitem{ARXIV_2017_Tran}
D.~Tran, J.~Ray, Z.~Shou, S.-F. Chang, and M.~Paluri, ``Convnet architecture
  search for spatiotemporal feature learning,'' \emph{CoRR}, vol.
  abs/1708.05038, 2017.

\bibitem{CVPR_2017_Carreira}
J.~Carreira and A.~Zisserman, ``Quo vadis, action recognition? {A} new model
  and the kinetics dataset,'' in \emph{{IEEE} International Conference on
  Computer Vision and Pattern Recognition (CVPR'17)}, 2017, pp. 4724--4733.

\bibitem{ARXIV_2017_Kay}
W.~Kay, J.~Carreira, K.~Simonyan, B.~Zhang, C.~Hillier, S.~Vijayanarasimhan,
  F.~Viola, T.~Green, T.~Back, P.~Natsev, M.~Suleyman, and A.~Zisserman, ``The
  kinetics human action video dataset,'' \emph{CoRR}, vol. abs/1705.06950,
  2017.

\end{thebibliography}
%\begin{thebibliography}{1}

%\end{thebibliography}


\end{document}